\documentclass[runninghead]{llncs}

\usepackage{algorithmicx}
\usepackage[noend]{algpseudocode}
\usepackage{graphicx}
\usepackage{multirow}
\begin{document}
\title{\.{I}NSANSIZ ARA\c{C}LARLA D\"UZLEMSEL OLMAYAN ALANLARIN TARANMASI}
\titlerunning{\.{I}nsans{\i}z Ara\c{c}larla D\"uzlemsel Olmayan Alanlar{\i}n Taramas{\i}}
%
\author{\c{C}aglar Seylan\inst{1} \and
\"Ozg\"ur Sayg{\i}n Bican\inst{1} \and Fatih Semiz\inst{1}}
\authorrunning{C. Seylan et. al}
%
\institute{Middle East Technical University, Ankara, Turkey}
\maketitle              

\begin{abstract}
\setlength{\parindent}{0pt}
\textbf{\"{O}z}\par
G\"{u}n\"{u}m\"{u}zde insans{\i}z ara\c{c}larla alan taramas{\i}, yani bir alan{\i}n t\"{u}m\"{u}n\"{u}n veya bir k{\i}sm{\i}n{\i}n insans{\i}z ara\c{c}larla en az efor ile dola\c{s}{\i}lmas{\i}, alan taramas{\i}na duyulan ihtiya\c{c} ve insans{\i}z ara\c{c}lar{\i}n kullan{\i}m{\i}n{\i}n artmas{\i}yla beraber h{\i}zla \"{o}nem kazanmaktad{\i}r. \.{I}nsans{\i}z ara\c{c}larla alan taramas{\i}n{\i}n, \.{I}HA'lar (\.{I}nsans{\i}z Hava Ara\c{c}lar{\i}) ile bir alanda ke\c{s}if yapmaktan robotlar ile may{\i}nl{\i} arazilerin may{\i}nlardan ar{\i}nd{\i}r{\i}lmas{\i}na, b\"{u}y\"{u}k al{\i}\c{s}veri\c{s} merkezlerinde yerlerin robotlarla temizlenmesinden b\"{u}y\"{u}k arazilerde \c{c}im bi\c{c}meye kadar pek \c{c}ok uygulamas{\i} mevcuttur. Problemin tek ara\c{c}la alan taramas{\i}, birden fazla ara\c{c}la alan taramas{\i}, \c{c}evrimi\c{c}i (arazinin nas{\i}l oldu\u{g}u daha \"{o}nceden bilinmiyorsa) alan taramas{\i} gibi pek \c{c}ok versiyonu mevcuttur. Ayr{\i}ca arazide \c{c}e\c{s}itli b\"{u}y\"{u}kl\"{u}klerde engeller de bulunabilmektedir. Do\u{g}al olarak, bu problem \"{u}zerinde bir\c{c}ok ara\c{s}t{\i}rmac{\i} \c{c}al{\i}\c{s}maktad{\i}r ve g\"{u}n\"{u}m\"{u}ze kadar pek \c{c}ok \c{c}al{\i}\c{s}ma yap{\i}lm{\i}\c{s}t{\i}r. Bu \c{c}al{\i}\c{s}\\-malar{\i}n \c{c}ok b\"{u}y\"{u}k bir k{\i}sm{\i}, MKA (Minimum Kapsama A\u{g}ac{\i}) yakla\c{s}{\i}-\\m{\i}n{\i} kullanmaktad{\i}r. Bu
yakla\c{s}{\i}mda temel olarak, d\"{u}zlemsel bir arazi, arac{\i}n g\"{o}r\"{u}\c{s} alan{\i}na g\"{o}re e\c{s} b\"{u}y\"{u}kl\"{u}kte karelere b\"{o}l\"{u}nmekte ve bu karele-\\rin merkezleri birer d\"{u}\u{g}\"{u}m olarak kabul edilerek araziden kenarlar{\i} birim a\u{g}{\i}rl{\i}kta olan bir \c{c}izge elde edilmektedir. 
Sonra bu \c{c}izgenin MKA's{\i} bulunup bu MKA'n{\i}n etraf{\i} ara\c{c}lar taraf{\i}ndan turlanmaktad{\i}r. Bizim \"{o}nerdi\u{g}imiz metot d\"{u}zlemsel olmayan arazilerin de insans{\i}z ara\c{c}lar ile taranmas{\i}na \c{c}\"{o}z\"{u}m getirmektedir. Biz de \c{c}\"{o}z\"{u}mde MKA yakla\c{s}{\i}m{\i}n{\i} temel ald{\i}k, ancak arazi d\"{u}zlemsel olmad{\i}\u{g}{\i} i\c{c}in \c{c}izgedeki kenarlara birim a\u{g}{\i}rl{\i}k vermek yerine iki kare aras{\i}ndaki e\u{g}ime ba\u{g}l{\i} a\u{g}{\i}rl{\i}klar verdik. Bu yakla\c{s}{\i}m ile ayn{\i} zamanda \"{o}zellikle \.{I}HA'lar i\c{c}in r\"{u}zgar{\i}n \c{s}iddeti ve y\"{o}n\"{u} de hesaba kat{\i}larak bir rota elde edilip alan taramas{\i} yap{\i}labilir.\\

\textbf{Anahtar Kelimeler:} alan taramas{\i}, \c{c}izge teorisi, \.{I}HA (\.{I}nsans{\i}z Hava Arac{\i}), MKA
(Minimum Kapsama A\u{g}ac{\i}), rota planlama\\

\textbf{Abstract}\par 
The importance of area coverage with unmanned vehicles, in other words, traveling an
area with an unmanned vehicle such as a robot or an UAV (Unmanned Aerial Vehicle)
completely or partially with minimum cost, is increasing with the increase in usage of
such vehicles today. Area coverage with unmanned vehicles are used today in
exploration of an area with a UAV, sweeping mines with robots, cleaning ground with
robots in large shopping malls, mowing lawn in a large area etc. The problem has
versions such as area coverage with a single unmanned vehicle, area coverage with
multiple unmanned vehicles, on-line area coverage (The map of the area that will be
covered is not known before starting the coverage) with unmanned vehicles etc. In
addition, the area may have obstacles that the vehicles cannot move over. Naturally,
many researches are working on the problem and a lot of researches have been done on
the problem until today. Spanning tree coverage is one of the major approaches to the
problem. In this approach, at the basic level, the planar area is divided into identical
squares according to range of sight of the vehicle, and centers of these squares are
assumed to be vertexes of a graph. The vertexes of this graph are connected with the
edges with unit costs and after finding MST (Minimum Spanning Tree) of the graph, the
vehicle strolls around the spanning tree. The method we propose suggests a way to
cover a non-planar area with unmanned vehicles. The method we propose also takes
advantage of spanning tree coverage approach, but instead of assigning unit costs to the
edges, we assigned a weight to each edge using slopes between vertexes connected with
those edges. We have gotten noticeably better results than the results we got when we
did not consider the slope between two squares and used classical spanning tree
approach.\\

\end{abstract}
\section{G\.{I}R\.{I}\c{S}}
G\"{u}n\"{u}m\"{u}zde insans{\i}z ara\c{c}lar i\c{c}in en \"{o}nemli problemlerden birisi alan taramas{\i}d{\i}r. Alan
taramas{\i} bir arazinin her yerini dola\c{s}mak olarak tan{\i}mlanabilir. Alan taramas{\i} yap{\i}l{\i}rken
g\"{o}z \"{o}n\"{u}nde bulundurulan en \"{o}nemli k{\i}staslardan bir tanesi bu taramay{\i} olabildi\u{g}ince
verimli yapmakt{\i}r. Bu amaca ula\c{s}mak i\c{c}in, bir yandan taranan alan maksimize edilmeye
\c{c}al{\i}\c{s}{\i}l{\i}rken, bir yandan da harcanan zaman ve yak{\i}t minimize edilmeye \c{c}al{\i}\c{s}{\i}l{\i}r.
Arazileri verimli bir \c{s}ekilde taramak i\c{c}in en s{\i}k kullan{\i}lan yakla\c{s}{\i}mlardan bir tanesi
kapsama a\u{g}ac{\i} y\"{o}ntemidir. Kapsama a\u{g}ac{\i}n{\i}n tan{\i}m{\i} \c{s}u \c{s}ekildedir:

G ba\u{g}l{\i} bir \c{c}izge olsun. Hem a\u{g}a\c{c} olan hem de G’nin t\"{u}m d\"{u}\u{g}\"{u}m noktalar{\i}n{\i} kapsayan
bir alt \c{c}izgeye kapsama a\u{g}ac{\i} denir.

Problemin tek ara\c{c}la alan taramas{\i}, birden fazla ara\c{c}la alan taramas{\i}, \c{c}evrimi\c{c}i
(Taranacak alan{\i}n haritas{\i} daha \"{o}nceden bilinmiyorsa) alan taramas{\i}, \c{c}evrimd{\i}\c{s}{\i} alan
taramas{\i} veya engel bulunan arazilerin taranmas{\i}, engel bulunmayan arazilerin
taranmas{\i} gibi \c{c}e\c{s}itli varyasyonlar{\i} mevcuttur.

G\"{u}n\"{u}m\"{u}zde alan taramas{\i}n{\i}n \.{I}HA'lar ile bir alanda ke\c{s}if yapmaktan robotlar ile may{\i}nl{\i}
arazilerin may{\i}nlardan ar{\i}nd{\i}r{\i}lmas{\i}na, b\"{u}y\"{u}k al{\i}\c{s}veri\c{s} merkezlerinde yerlerin robotlarla
temizlenmesinden b\"{u}y\"{u}k arazilerde \c{c}im bi\c{c}meye kadar pek \c{c}ok uygulamas{\i} mevcuttur
\cite{yedi}, \cite{dort}, \cite{sekiz}.

Bu konuda bug\"{u}ne kadar yap{\i}lan \c{c}al{\i}\c{s}malar{\i}n \c{c}ok b\"{u}y\"{u}k bir k{\i}sm{\i} taranacak arazileri
d\"{u}zlemsel olarak ele alm{\i}\c{s}t{\i}r. Bu \c{c}al{\i}\c{s}man{\i}n amac{\i}, d\"{u}zlemsel arazilerde alan taramas{\i}
i\c{c}in geli\c{s}tirilen mevcut yakla\c{s}{\i}mlar \"{u}zerine geli\c{s}tirmeler yap{\i}larak d\"{u}zlemsel olmayan
arazilerin de verimli olarak taranabilece\u{g}i bir yakla\c{s}{\i}m t\"{u}ret-\\mektir.
\section{\.{I}LG\.{I}L\.{I} \c{C}ALI\c{S}MALAR}
Alan taramas{\i}nda, en \"{o}nemli problemlerden bir tanesi arazinin bilgisayar ortam{\i}nda
nas{\i}l g\"{o}sterilece\u{g}idir. Daha a\c{c}{\i}k s\"{o}ylemek gerekirse problem, s\"{u}rekli ger\c{c}ek d\"{u}nyan{\i}n
kesikli bilgisayar d\"{u}nyas{\i}nda nas{\i}l g\"{o}sterilebilece\u{g}idir. \c{C}\"{o}z\"{u}mler-den bir tanesi A. Elfes
taraf{\i}ndan \"{o}nerilen araziye, araziyi e\c{s} h\"{u}crelere b\"{o}lerek \\yak{\i}nsamakt{\i}r \cite{bir}. Bilinmeyen
ve yap{\i}land{\i}r{\i}lmam{\i}\c{s} alanlar i\c{c}in, Elfes sonar tabanl{\i} ger\c{c}ek d\"{u}nya harita ve navigasyon
sistemi \"{o}nermi\c{s}tir. B\"{o}ylece araziyi bilgisayar ortam{\i}nda g\"{o}sterebilmek i\c{c}in bir yakla\c{s}{\i}m
bulmu\c{s}tur. Bu yakla\c{s}{\i}mda arazi, e\c{s} h\"{u}crelerden olu\c{s}an iki boyutlu bir dizi olarak
g\"{o}sterilmi\c{s}tir. H\"{u}cre, h\"{u}creye bir ara\c{c} girince veya arac{\i}n alg{\i}lay{\i}c{\i}s{\i} h\"{u}creyi
alg{\i}lay{\i}nca gezilmi\c{s} olarak kabul edilir. 2 boyutlu dizideki t\"{u}m h\"{u}creler gezildi\u{g}inde
alan taranm{\i}\c{s} olur.
	
\.{I}nsans{\i}z ara\c{c}larla alan taramas{\i}na di\u{g}er h\"{u}cre tabanl{\i} bir yakla\c{s}{\i}m ise Y. Gabriely ve E.
Rimon taraf{\i}ndan \"{o}nerilen kapsama a\u{g}ac{\i} yakla\c{s}{\i}m{\i}d{\i}r \cite{iki}. Bu yakla\c{s}{\i}mda araziye, arazi
e\c{s} h\"{u}crelere b\"{o}l\"{u}n\"{u}p daha sonra bu h\"{u}crelerden bir \c{c}izge elde edilerek yak{\i}nsanm{\i}\c{s}t{\i}r.
Daha sonra bu \c{c}izgenin kapsama a\u{g}ac{\i} bulunur ve etraf{\i} bir ya da birden fazla ara\c{c}la
dola\c{s}{\i}l{\i}r. Dola\c{s}ma bitti\u{g}inde alan taranm{\i}\c{s} olur.

S. Hert, S. Tiwari ve V. Lumelsky otonom su alt{\i} ara\c{c}lar{\i}na 3 boyutlu su alt{\i} alanlar{\i}n{\i}n
taranmas{\i} i\c{c}in rota plan{\i} yapan k{\i}smi kesikli bir algoritma \"{o}nermi\c{s}tir \cite{uc}. \"{O}nerilen
algoritma \c{c}evrimi\c{c}idir (Taranacak olan su alt{\i}ndaki alan tarama ba\c{s}lamadan \"{o}nce
bilinmemektedir). Bu yakla\c{s}{\i}mdaki h\"{u}creler tamamen sabit de\u{g}ildir, enleri sabit fakat
boylar{\i} de\u{g}i\c{s}kendir. Bu algoritma ile basit ba\u{g}lant{\i}l{\i} ya da basit olmayan ba\u{g}lant{\i}l{\i}
alanlar taranabilmektedir. Algoritman{\i}n \"{o}zyinelemeli bir do\u{g}as{\i} vard{\i}r ve ara\c{c} alan{\i} paralel d\"{u}z \c{c}izgiler boyunca zigzag hareketi yaparak tarar. Alan adalar ve koylar i\c{c}erebilir. Koylara girerken ya da \c{c}{\i}karken koylar{\i}n i\c{c}indeki baz{\i} alanlar birden fazla
taranabilir ya da baz{\i} alanlar hi\c{c} taranmayabilir (Bu t\"{u}r koylara sapt{\i}r{\i}c{\i} koy denir). Bu
t\"{u}r koylar{\i} taramak i\c{c}in \"{o}zel yakla\c{s}{\i}mlara gerek vard{\i}r. Alan ada i\c{c}erdi\u{g}i zaman, yani
alan basit ba\u{g}lant{\i}l{\i} de\u{g}ilse, ayn{\i} prosed\"{u}rler \"{u}zerinde k\"{u}\c{c}\"{u}k de\u{g}i\c{s}iklikler yap{\i}larak
alan{\i}n tamamen taranmas{\i} sa\u{g}lan{\i}r.

Alan{\i}n h\"{u}crelere ayr{\i}lmas{\i} bir yak{\i}nsama olmak zorunda de\u{g}ildir, tam da olabilir. Alan
tam olarak h\"{u}crelere ayr{\i}l{\i}rken, \"{o}yle farkl{\i} b\"{o}lgelere ayr{\i}l{\i}r ki bu b\"{o}lgelerin birle\c{s}imi
alan{\i} tam olarak verir \cite{dort}. Tam olarak h\"{u}crelere ay{\i}rma tekniklerinden bir tanesi alan{\i}
yamuksal \c{s}ekilde ay{\i}rmakt{\i}r. J. R. VanderHeide ve N. S. V. Rao yamuksal \c{s}ekilde alan
ay{\i}rma tabanl{\i} \c{c}evrimi\c{c}i bir alan taramas{\i} metodu \"{o}nermi\c{s}tir \cite{bes}. Temel olarak, insans{\i}z
ara\c{c} alan hakk{\i}nda bilgi toplama faz{\i}ndan sonra alan{\i} dikd\"{o}rtgenlere ay{\i}r{\i}r ve her
dikd\"{o}rtgen i\c{c}in ayr{\i} bir rota planlamas{\i} yapar. Daha sonra her dikd\"{o}rtgen i\c{c}in bulunan
rotalar{\i} birle\c{s}tirir ve tek bir rota elde eder.
\section{\"{O}NER\.{I}LEN YAKLA\c{S}IM}
Bug\"{u}ne kadar \"{o}nerilen metotlar genelde d\"{u}z alanlarda alan taramas{\i} yapmak i\c{c}indir. Biz
bu yaz{\i}da d\"{u}zlemsel olmayan, engebeli y\"{u}zeyleri taramak i\c{c}in, insans{\i}z ara\c{c}lara rota
planlamas{\i} yapan bir yakla\c{s}{\i}m sunuyoruz. Bizim yakla\c{s}{\i}m{\i}m{\i}z da temel olarak Y.
Gabriely ve E. Rimon’un yakla\c{s}{\i}m{\i} \cite{iki} gibi Kapsama A\u{g}ac{\i} Taramas{\i} y\"{o}ntemini
kullanmaktad{\i}r. Ancak bu y\"{o}ntemde, kapsama a\u{g}ac{\i} olu\c{s}-turulurken, araziden elde
edilen \c{c}izgenin kenarlar{\i}na birim a\u{g}{\i}rl{\i}k yerine kom\c{s}u iki d\"{u}\u{g}\"{u}m aras{\i}ndaki e\u{g}ime ba\u{g}l{\i}
bir a\u{g}{\i}rl{\i}k atan{\i}r. Daha sonra olu\c{s}an \c{c}izgenin minimum kapsama a\u{g}ac{\i} bulunarak bu
a\u{g}ac{\i}n etraf{\i} dola\c{s}{\i}l{\i}r.

Arazi, bilgisayarda her bir eleman{\i} o noktan{\i}n y\"{u}ksekli\u{g}ini (Belirli bir yere g\"{o}re)
belirten bir N x M’lik matris ile g\"{o}sterilir. \c{S}ekil 1’de bu matrise bir \"{o}rnek
g\"{o}r\"{u}lmektedir.

\begin{figure}[!h]        
  \begin{center}    
 \includegraphics[scale=0.25]{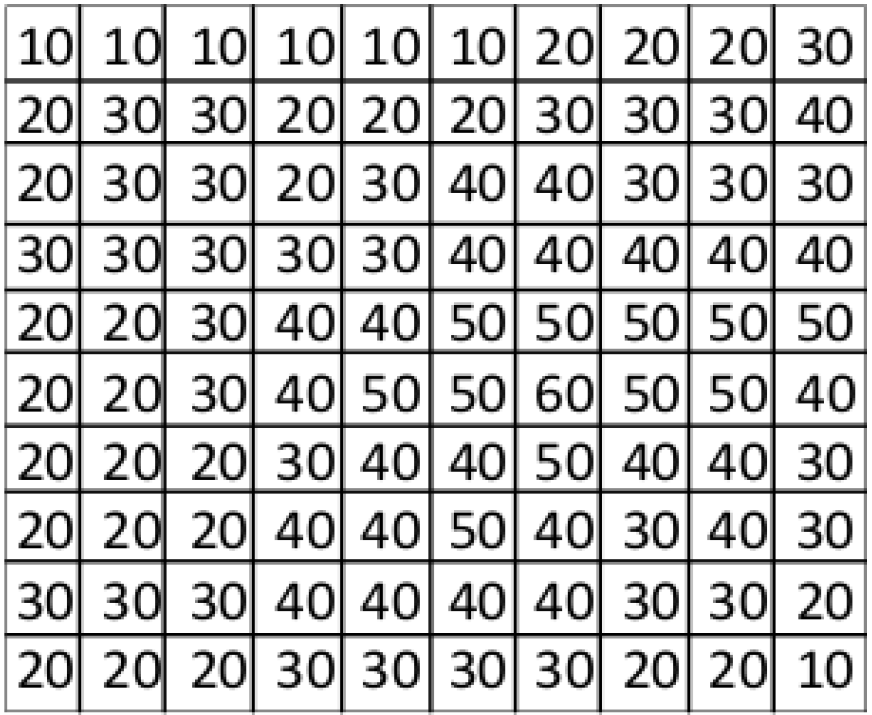}    
    \caption{Arazi i\c{c}in \"{o}rnek matris.}    
    \label{Figure:sekil1}   
  \end{center}     
\end{figure}

\.{I}nsans{\i}z araca ba\u{g}l{\i} kameran{\i}n g\"{o}r\"{u}\c{s} a\c{c}{\i}s{\i}n{\i}n, ya da daha genel bir s\"{o}ylemle araca ba\u{g}l{\i}
cihaz{\i}n kapsama alan{\i}n{\i}n, matriste bir karelik alana denk geldi\u{g}i farz edilmektedir. Bu
matris N. Hazon’{\i}n yakla\c{s}{\i}m{\i}ndaki gibi her kenar{\i} iki kare olan (Daha genel bir
s\"{o}ylemle arac{\i}n kapsama alan{\i} \c{c}ap{\i}n{\i}n iki kat{\i} uzunlu\u{g}unda) daha b\"{u}y\"{u}k karelere ayr{\i}l{\i}r
\cite{alti} (\c{S}ekil 2).

\begin{figure}[!h]        
  \begin{center}    
 \includegraphics[scale=0.25]{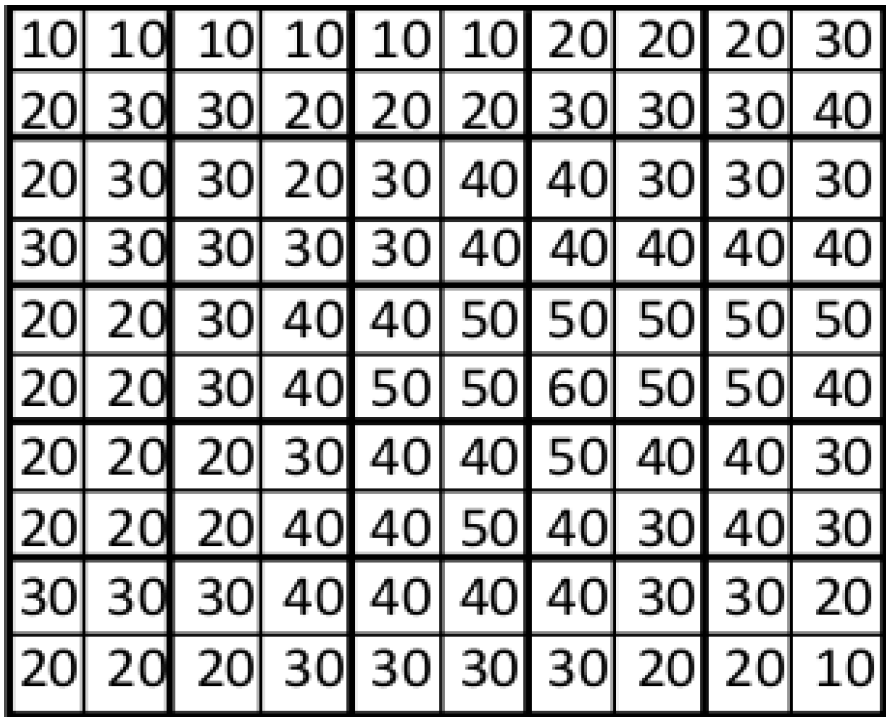}    
    \caption{Alt karelere ayr{\i}lm{\i}\c{s} matris.}    
    \label{Figure:sekil2}   
  \end{center}     
\end{figure}

Bu matristen (N/2) x (M/2)’lik yeni bir matris \"{u}retilir. Bu matrisin her bir eleman{\i} eski
matrisin 2 x 2’lik karelerinin i\c{c}indeki elemanlara (\c{S}ekil 2’de kal{\i}n s{\i}n{\i}rlarla belirtilen)
denk gelmektedir. Yeni matristeki her bir eleman da eski matristeki gibi y\"{u}kseklik
belirtmektedir. Bu y\"{u}kseklikler eski matristeki 2 x 2’lik karelerdeki 4 eleman{\i}n
aritmetik ortalamas{\i} al{\i}narak bulunur ve ger\c{c}ek araziye yak{\i}nsama yap{\i}l{\i}r. Yani, bir 2 x
2’lik kare i\c{c}indeki elemanlar H1, H2, H3, H4 ise yeni matriste bu kareye denk gelen
eleman{\i}n de\u{g}eri Hyeni, bu d\"{o}rt eleman{\i}n aritmetik ortalamas{\i}d{\i}r (\c{S}ekil 3).

\begin{figure}[!h]        
  \begin{center}    
 \includegraphics[scale=0.22]{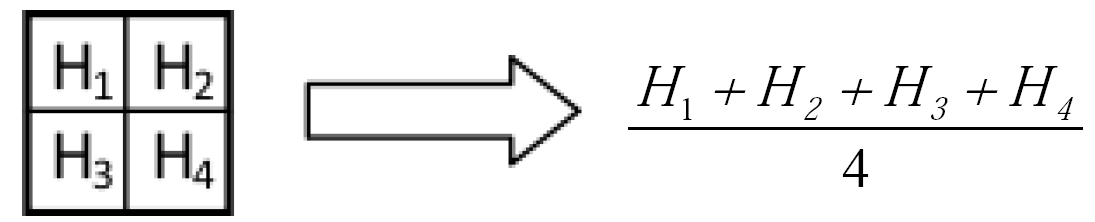}    
    \caption{Alt matristeki d\"{o}rt eleman{\i}n aritmetik ortalamas{\i}.}    
    \label{Figure:sekil3}   
  \end{center}     
\end{figure}

Yeni matristeki her bir h\"{u}cre ayn{\i} zamanda bir d\"{u}\u{g}\"{u}m noktas{\i}d{\i}r. Her d\"{u}\u{g}\"{u}m noktas{\i}
ile, kendisine kar\c{s}{\i}l{\i}k gelen h\"{u}crenin kom\c{s}ular{\i} kenarlar arac{\i}l{\i}\u{g}{\i}yla ba\u{g}l{\i}d{\i}r (Her bir
h\"{u}cre, kendisinin sa\u{g}{\i}ndaki, solundaki, \"{u}st\"{u}ndeki ve alt{\i}ndaki h\"{u}cre ile kom\c{s}udur,
\c{c}apraz h\"{u}creler kom\c{s}udan say{\i}lmaz).
Kenarlar{\i}n a\u{g}{\i}rl{\i}klar{\i}, ba\u{g}lad{\i}klar{\i} d\"{u}\u{g}\"{u}m noktalar{\i} aras{\i}ndaki e\u{g}ime ba\u{g}l{\i} bir
fonksiyon arac{\i}l{\i}\u{g}{\i}yla bulunur. Bu fonksiyon, d\"{u}\u{g}\"{u}m noktalar{\i} aras{\i}ndaki yatay uzakl{\i}k
d, y\"{u}kseklikleri aras{\i}ndaki fark h ise, do\u{g}rudan aradaki uzakl{\i}\u{g}{\i} veren Pisagor ba\u{g}{\i}nt{\i}s{\i},
yani $\sqrt{d^2 + h^2}$, \"{o}zellikle \.{I}HA'lar i\c{c}in r\"{u}zgar{\i}n y\"{o}n\"{u} ve \c{s}iddetine ba\u{g}l{\i} bir ba\u{g}{\i}nt{\i}, ya da
amaca g\"{o}re \c{c}ok daha karma\c{s}{\i}k bir ba\u{g}{\i}nt{\i} olabilir.

\c{C}izge bu \c{s}ekilde olu\c{s}turulduktan sonra minimum kapsama a\u{g}ac{\i} bulunur ve bu kapsama
a\u{g}ac{\i}n{\i}n etraf{\i} Y. Gabriely ve E. Rimon’un yapt{\i}\u{g}{\i} yakla\c{s}{\i}mdaki gibi ara\c{c}lar taraf{\i}ndan
turlan{\i}r \cite{iki}. Y\"{u}kseklik matrisi \c{S}ekil 1’deki gibi olan matriste, do\u{g}rudan Pisagor
ba\u{g}{\i}nt{\i}s{\i}na g\"{o}re bulunmu\c{s} \c{c}izgenin, bu yakla\c{s}{\i}ma g\"{o}re turlan{\i}rken ortaya \c{c}{\i}kan rota
\c{s}ekil 4’te g\"{o}sterilmi\c{s}tir.
\section{TEST SONU\c{C}LARI}
Testlerde 15 tane girdi kullan{\i}ld{\i}. Girdiler, rastgele \"{u}retilmi\c{s} araziler olup 250x250
matris \c{s}eklinde g\"{o}sterilmi\c{s}tir. Girdilerdeki arazilere en fazla 30 adet engel koyul-mu\c{s}tur.
Testlerde bizim \"{o}nerdi\u{g}imiz yakla\c{s}{\i}m ile \c{s}imdiye kadar d\"{u}zlemsel arazileri taramada
kullan{\i}lan do\u{g}rudan kapsama a\u{g}ac{\i} yakla\c{s}{\i}m{\i}n{\i} bu 15 girdiyi her iki yakla\c{s}{\i}m i\c{c}in de
kullanarak kar\c{s}{\i}la\c{s}t{\i}rd{\i}k. Bunu yaparken her birinin ortalama e\u{g}imi farkl{\i} olan girdiler
denedik.

Ortalama e\u{g}im, b\"{u}t\"{u}n kom\c{s}u h\"{u}cre ikililerinden elde edilen e\u{g}imlerin topla-m{\i}n{\i}n
\c{c}izgedeki kenar say{\i}s{\i}na b\"{o}l\"{u}m\"{u}nden elde edilmi\c{s}tir. \.{I}ki h\"{u}cre aras{\i}ndaki e\u{g}im, iki
h\"{u}crenin ortalama y\"{u}kseklikleri aras{\i}ndaki fark{\i}n bu h\"{u}creler aras{\i}ndaki yatay uzakl{\i}\u{g}a
b\"{o}l\"{u}m\"{u} olarak tan{\i}mlanm{\i}\c{s}t{\i}r.

Testlerde metot, \c{c}izgedeki kenarlara a\u{g}{\i}rl{\i}k atamada kullan{\i}lan iki farkl{\i} a\u{g}{\i}rl{\i}k
fonksiyonu kullan{\i}larak denenmi\c{s}tir. Bu fonksiyonlardan ilki, kenarlara do\u{g}rudan
Pisagor ba\u{g}{\i}nt{\i}s{\i} kullan{\i}larak bulunan, h\"{u}creler aras{\i}ndaki do\u{g}rusal uzakl{\i}\u{g}{\i}
bulmaktad{\i}r. Di\u{g}eri ise yine ilk fonksiyon gibi Pisagor ba\u{g}{\i}nt{\i}s{\i}n{\i} kullanarak h\"{u}creler
aras{\i}ndaki do\u{g}rusal uzakl{\i}\u{g}{\i} bulmaktad{\i}r. Ancak bunu do\u{g}rudan kenarlar{\i}n a\u{g}{\i}rl{\i}\u{g}{\i}
olarak kullanmak yerine buna, artan e\u{g}im ile beraber artan bir ceza puan{\i} eklemekdedir.
Fonksiyonlar ve test sonu\c{c}lar{\i} daha detayl{\i} olarak a\c{s}a\u{g}{\i}da a\c{c}{\i}klanm{\i}\c{s}t{\i}r.

\subsection{Do\u{g}rudan pisagor ba\u{g}{\i}nt{\i}s{\i}n{\i} kullanan a\u{g}{\i}rl{\i}k fonksiyonu}
Ortalama y\"{u}kseklikleri h1 ve h2 olan, aralar{\i}ndaki yatay uzakl{\i}k ise d olan h\"{u}crelere
kar\c{s}{\i}l{\i}k gelen d\"{u}\u{g}\"{u}mler aras{\i}ndaki kenarlara,
\begin{equation}
 w= \sqrt{|h_1-h_2|^2 - d^2}
\end{equation}

fonksiyonuna g\"{o}re a\u{g}{\i}rl{\i}k atanm{\i}\c{s}t{\i}r. Girdi numaras{\i}na g\"{o}re ortalama e\u{g}im, kapsama
a\u{g}ac{\i} yakla\c{s}{\i}m{\i}na g\"{o}re bulunan a\u{g}a\c{c}taki kenarlar{\i}n a\u{g}{\i}rl{\i}klar{\i} toplam{\i} ve \"{o}nerdi\u{g}imiz
metoda g\"{o}re (Herhangi bir kapsama a\u{g}ac{\i}ndan ziyade minimum kapsama a\u{g}ac{\i}na g\"{o}re)
bulunan a\u{g}a\c{c}taki kenarlar{\i}n a\u{g}{\i}rl{\i}klar{\i} toplam{\i} \c{c}izelge 1’de g\"{o}sterilmi\c{s}tir. E\u{g}ime ba\u{g}l{\i}
olarak de\u{g}i\c{s}en kenar a\u{g}{\i}rl{\i}klar{\i} toplamlar{\i} \c{c}izelge 2’de g\"{o}sterilmi\c{s}tir (Dikey eksen
a\u{g}{\i}rl{\i}klar toplam{\i}na, yatay eksen ortalama e\u{g}ime kar\c{s}{\i}l{\i}k gelmektedir.).

\begin{figure}[!h]        
  \begin{center}    
 \includegraphics[scale=0.45]{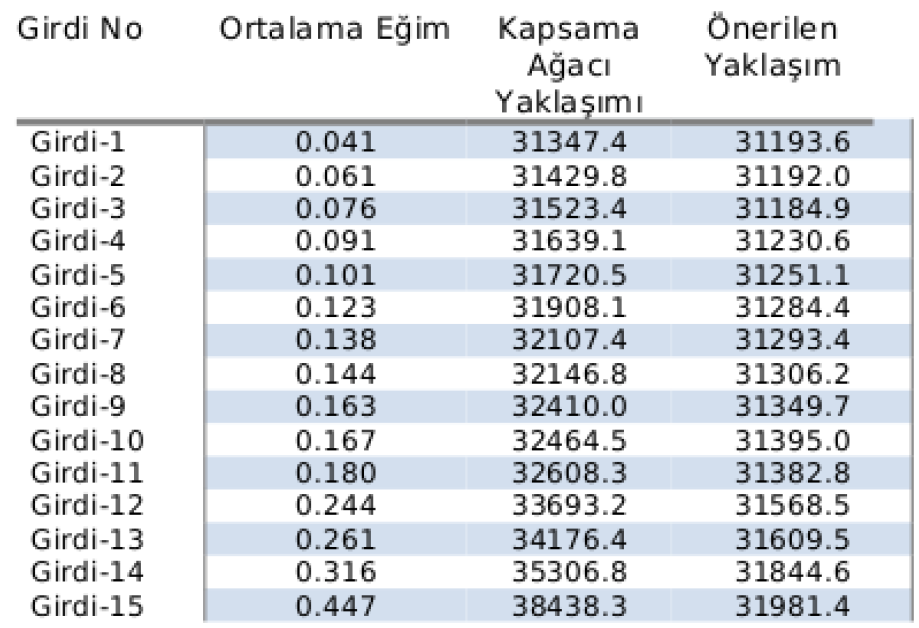}    
    \caption{\.{I}ki metodun denklem (1)'e g\"{o}re kar\c{s}{\i}la\c{s}t{\i}r{\i}lmas{\i}.}    
    \label{Figure:sekil4}   
  \end{center}     
\end{figure}

\begin{figure}[!h]        
  \begin{center}    
 \includegraphics[scale=0.35]{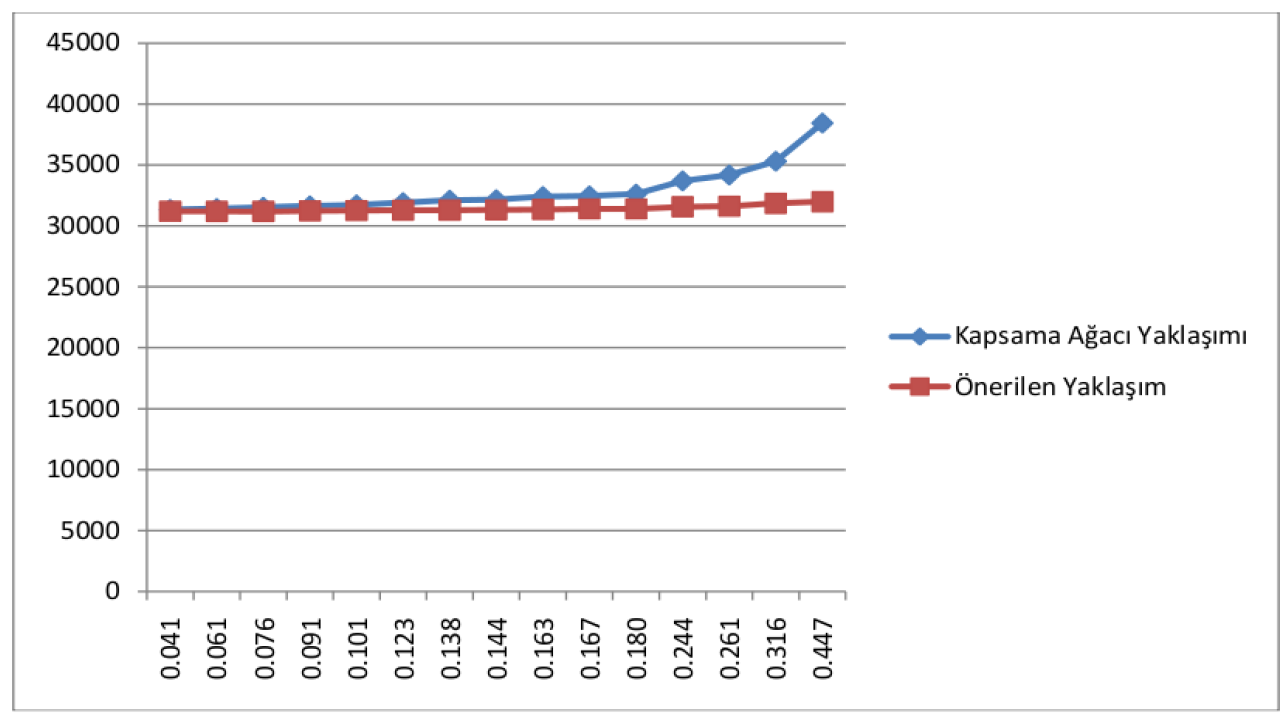}    
    \caption{\.{I}ki metodun denklem (1)'e g\"{o}re kar\c{s}{\i}la\c{s}t{\i}r{\i}lmas{\i}.}    
    \label{Figure:sekil5}   
  \end{center}     
\end{figure}

\subsection{Ek olarak e\u{g}ime ba\u{g}l{\i} ceza puan{\i} da kullanan a\u{g}{\i}rl{\i}k fonksiyonu}
Ortalama y\"{u}kseklikleri h1 ve h2 olan, aralar{\i}ndaki yatay uzakl{\i}k ise d olan h\"{u}crelere
kar\c{s}{\i}l{\i}k gelen d\"{u}\u{g}\"{u}mler aras{\i}ndaki kenarlara,
\begin{equation}
 w= \sqrt{|h_1-h_2|^2 - d^2} * (1 + \frac{|h_1-h_2|}{d})
\end{equation}

fonksiyonuna g\"{o}re a\u{g}{\i}rl{\i}k atanm{\i}\c{s}t{\i}r. Bu fonksiyon, ilk fonksiyona yine e\u{g}ime ba\u{g}l{\i} bir
ceza puan{\i} eklenerek elde edilmi\c{s}tir. Girdi numaras{\i}na g\"{o}re ortalama e\u{g}im, kapsama
a\u{g}ac{\i} yakla\c{s}{\i}m{\i}na g\"{o}re bulunan a\u{g}a\c{c}taki kenarlar{\i}n a\u{g}{\i}rl{\i}klar{\i} toplam{\i} ve \"{o}nerdi\u{g}imiz
metoda g\"{o}re (Herhangi bir kapsama a\u{g}ac{\i}ndan ziyade minimum kapsama a\u{g}ac{\i}na g\"{o}re)
bulunan a\u{g}a\c{c}taki kenarlar{\i}n a\u{g}{\i}rl{\i}klar{\i} toplam{\i} \c{c}izelge 3’de g\"{o}sterilmi\c{s}tir. E\u{g}ime ba\u{g}l{\i}
olarak de\u{g}i\c{s}en kenar a\u{g}{\i}rl{\i}klar{\i} toplamlar{\i} \c{c}izelge 4’de g\"{o}sterilmi\c{s}tir (Dikey eksen
a\u{g}{\i}rl{\i}klar toplam{\i}na, yatay eksen ortalama e\u{g}ime kar\c{s}{\i}l{\i}k gelmektedir.).

\begin{figure}[!h]        
  \begin{center}    
 \includegraphics[scale=0.35]{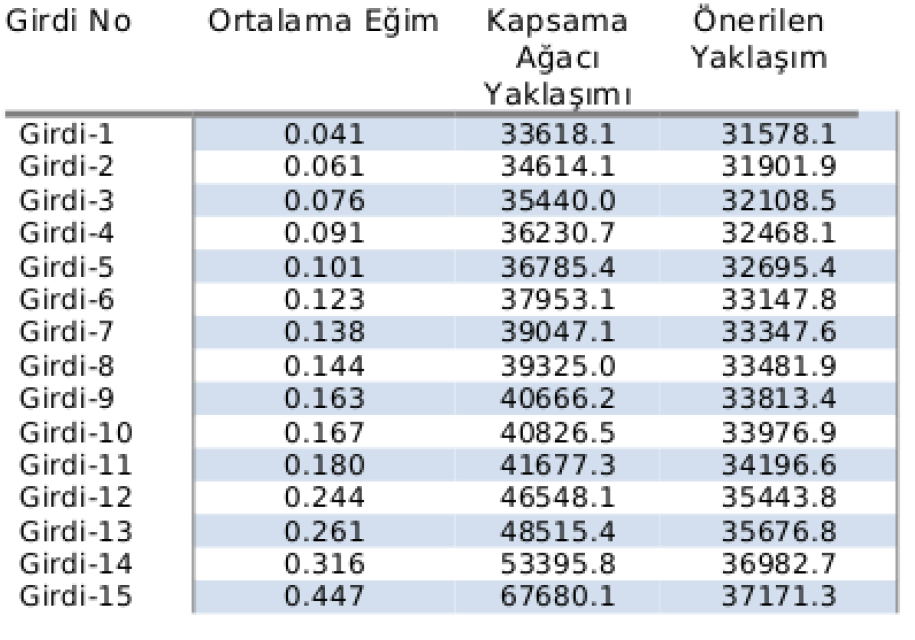}    
    \caption{\.{I}ki metodun denklem (2)'e g\"{o}re kar\c{s}{\i}la\c{s}t{\i}r{\i}lmas{\i}.}    
    \label{Figure:sekil6}   
  \end{center}     
\end{figure}

\begin{figure}[!h]        
  \begin{center}    
 \includegraphics[scale=0.35]{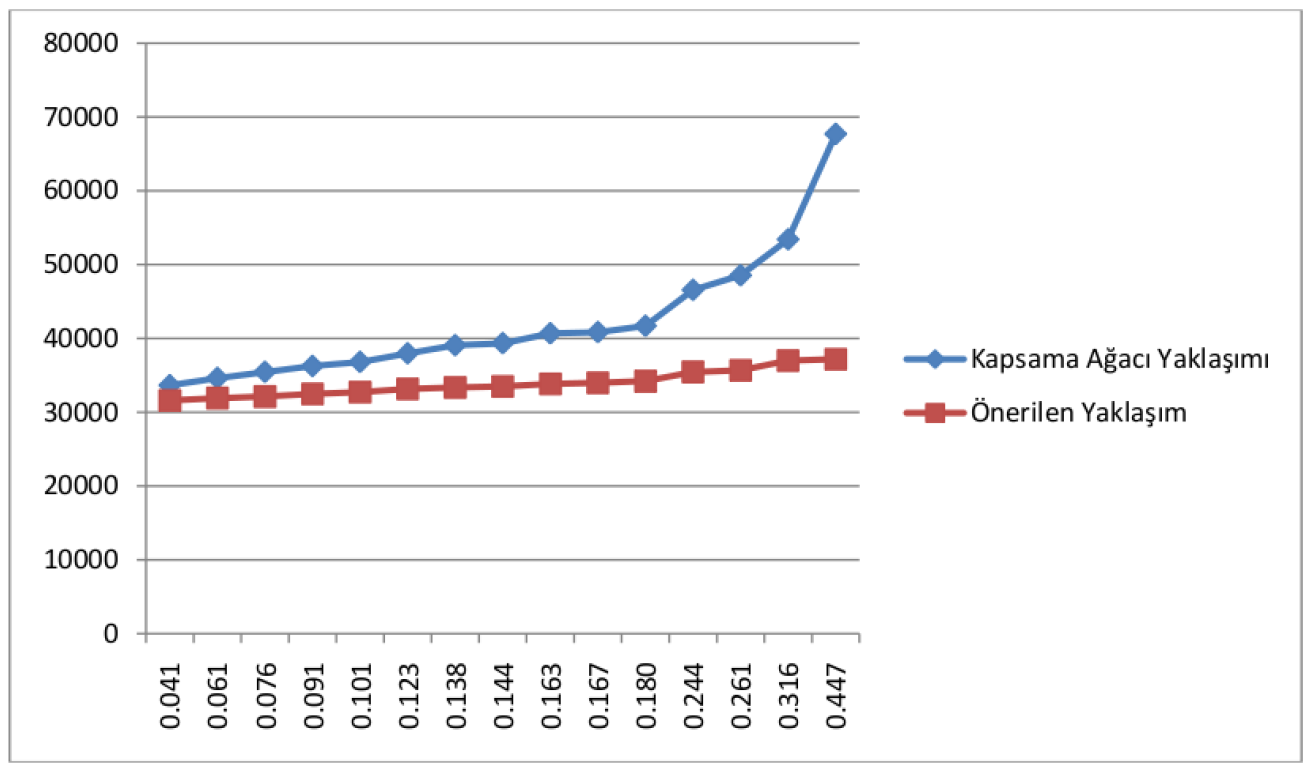}    
    \caption{\.{I}ki metodun denklem (2)'e g\"{o}re kar\c{s}{\i}la\c{s}t{\i}r{\i}lmas{\i}.}    
    \label{Figure:sekil7}   
  \end{center}     
\end{figure}

\subsection{Yorumlar}
\.{I}ki yakla\c{s}{\i}m do\u{g}rudan Pisagor ba\u{g}{\i}nt{\i}s{\i}n{\i} kullanan a\u{g}{\i}rl{\i}k fonksiyonu ile test
edildi\u{g}inde, \"{o}nerilen yakla\c{s}{\i}m di\u{g}er yakla\c{s}{\i}ma g\"{o}re biraz daha verimli rotalar
bulmaktad{\i}r. Arazinin ortalama e\u{g}imi artt{\i}k\c{c}a (\"{O}zellikle 0.5’e yakla\c{s}t{\i}k\c{c}a) aradaki fark
daha belirgin hale gelmektedir. Yakla\c{s}{\i}mlar Pisagor ba\u{g}{\i}nt{\i}s{\i}na ek olarak e\u{g}ime ba\u{g}l{\i}
ceza puan{\i} da kullanan a\u{g}{\i}rl{\i}k fonksiyonu ile test edildi\u{g}inde, yine \"{o}nerilen yakla\c{s}{\i}m
di\u{g}er yakla\c{s}{\i}ma g\"{o}re daha verimli rotalar bulmaktad{\i}r ancak aradaki fark bu sefer daha
belirgindir. Arazinin ortalama e\u{g}imi artt{\i}r{\i}ld{\i}k\c{c}a aradaki fark ilk a\u{g}{\i}rl{\i}k fonksiyonu
kullan{\i}ld{\i}\u{g}{\i} zamankine g\"{o}re \c{c}ok daha belirgin hale gelmektedir. \"{O}zellikle ortalama
e\u{g}im 0.5’e yakla\c{s}{\i}rken, \"{o}nerilen yakla\c{s}{\i}mda toplam a\u{g}{\i}rl{\i}k yava\c{s} bir art{\i}\c{s} g\"{o}stermi\c{s}tir,
ancak di\u{g}er yakla\c{s}{\i}mda toplam a\u{g}{\i}rl{\i}k neredeyse \"{u}stel bir bi\c{c}imde art{\i}\c{s} g\"{o}stermi\c{s}tir.
\section{SONU\c{C}}
Makalede d\"{u}zlemsel olmayan arazilerin insans{\i}z ara\c{c}larla verimli bir \c{s}ekilde taranmas{\i}
i\c{c}in yeni bir metot \"{o}nerilmi\c{s}tir. Bu metot, temel olarak, kapsama a\u{g}ac{\i} yakla\c{s}{\i}m{\i}nda
bulunan kenarlara birim a\u{g}{\i}rl{\i}k yerine e\u{g}ime ba\u{g}l{\i} bir fonksiyona g\"{o}re atanan a\u{g}{\i}rl{\i}klar{\i}
kullanmaktad{\i}r. Testlerde \"{o}nerilen metot kapsama a\u{g}ac{\i} metodu ile kar\c{s}{\i}la\c{s}t{\i}r{\i}lm{\i}\c{s}t{\i}r.
Genel olarak \"{o}nerilen metot di\u{g}er metoda g\"{o}re daha verimli sonu\c{c}lar verse de kenarlara
e\u{g}imi daha fazla hesaba katmak i\c{c}in atanan a\u{g}{\i}rl{\i}klarda e\u{g}imin uzakl{\i}\u{g}a g\"{o}re etkisi
daha da artt{\i}r{\i}ld{\i}\u{g}{\i}nda aradaki fark daha da artmaktad{\i}r.

D\"{u}zlemsel y\"{u}zeylerden elde edilen herhangi bir kapsama a\u{g}ac{\i} ayn{\i} zamanda minimum
kapsama a\u{g}ac{\i} olaca\u{g}{\i}ndan \"{o}nerilen yakla\c{s}{\i}mla arada rotan{\i}n verimi a\c{c}{\i}s{\i}ndan bir fark
olmaz. Ancak y\"{u}zey d\"{u}zlemsel de\u{g}ilse ortalama e\u{g}ime ba\u{g}l{\i} olarak minimum kapsama
a\u{g}ac{\i} herhangi bir kapsama a\u{g}ac{\i}na g\"{o}re daha verimli sonu\c{c}lar verir. Bu nedenle
herhangi bir kapsama a\u{g}ac{\i}n{\i}n etraf{\i} dola\c{s}{\i}lmaktansa minimum kapsama a\u{g}ac{\i}n{\i}n etraf{\i}
dola\c{s}{\i}l{\i}rsa, d\"{u}zlemsel olmayan arazilerde, tara-ma i\c{c}in daha verimli rotalar elde edilmi\c{s}
olur.
\section{TE\c{S}EKK\"{U}RLER}
Makalenin yazarlar{\i} olarak ODT\"{U}-TSK MODS\.{I}MMER’e bu makaleye deste\u{g}inden
dolay{\i} ve Do\c{c}. Dr. Veysi \.{I}\c{s}ler'e yapt{\i}\u{g}{\i} b\"{u}y\"{u}k katk{\i}lar ve yard{\i}mlardan dolay{\i}
te\c{s}ekk\"{u}r\"{u} bor\c{c} biliriz.

%
%
%
%

\bibliography{usmos.bib}
\end{document}